\documentclass[10pt,twocolumn,letterpaper]{article}

\usepackage{cvpr}              
\usepackage{verbatim}
\usepackage[ruled,vlined]{algorithm2e}

\definecolor{cvprblue}{rgb}{0.21,0.49,0.74}
\usepackage[pagebackref,breaklinks,colorlinks,allcolors=cvprblue]{hyperref}
\usepackage{multirow}
\usepackage{listings}
\usepackage[skins, breakable, many]{tcolorbox}
\usepackage{xcolor}
\tcbuselibrary{listings}

\newtcblisting{promptbox}{
  listing only,
  listing options={
    basicstyle=\ttfamily\small,
    breaklines=true
  },
  colback=white,
  colframe=black,
  arc=6pt,        
  boxrule=0.5pt,
  left=6pt,
  right=6pt,
  top=6pt,
  bottom=6pt
}

\def\confName{CVPR}
\def\confYear{2026}

\title{Devil is in Narrow Policy: Unleashing Exploration in  Driving VLA Models}

\author{
Canyu Chen$^{1,3}$\thanks{Equal contribution:\{chencanyu,guangbuaa\}@buaa.edu.cn} \quad
Yuguang Yang$^{2,3}$\footnotemark[1] \quad
Zhewen Tan$^{6}$ \quad
Yizhi Wang$^{7}$ \quad
Ruiyi Zhan$^{6}$ \\
Haiyan Liu$^{4}$ \quad
Xuanyao Mao$^{4}$ \quad
Jason Bao$^{4}$ \quad
Xinyue Tang$^{4}$ \\
Linlin Yang$^{5}$\thanks{Corresponding author: \texttt{lyang@cuc.edu.cn}, \texttt{sunbc1@lenovo.com}, \texttt{wangyan@air.tsinghua.edu.cn}} \quad
Bingchuan Sun$^{4}$\footnotemark[2] \quad
Yan Wang$^{3}$\footnotemark[2] \quad
Baochang Zhang$^{8}$\thanks{Project Lead} \\
\\
$^{1}$ National Superior College for Engineers, Beihang University \\
$^{2}$ School of Electronic Information Engineering, Beihang University\\
$^{3}$ Institute for AI Industry Research, Tsinghua University \quad 
$^{4}$ Lenovo Group Limited \\ $^{5}$ State Key Laboratory of Media Convergence and Communication, 
Communication University of China \\
 School of \{$^{6}$ Computer Science and Engineering, $^{7}$ Cyber Science and Technology, \\ $^{8}$ Artificial Intelligence \}, Beihang University
}

\begin{document}
\maketitle

\begin{abstract}
We identify a fundamental  \textbf{Narrow Policy} limitation undermining the performance of autonomous VLA models, where driving Imitation Learning (IL) tends to collapse exploration and limit the potential of subsequent Reinforcement Learning (RL) stages, which often saturate prematurely due to insufficient feedback diversity. Thereby, we propose {Curious-VLA}, a framework that alleviates the ``exploit–explore'' dilemma through a two-stage design.
During IL, we introduce a Feasible Trajectory Expansion (FTE) strategy to generate multiple physically valid trajectories and a step-wise normalized trajectory representation to adapt this diverse data. In the RL stage, we present {Adaptive Diversity-Aware Sampling (ADAS)} that prioritizes high-diversity samples and introduce {Spanning Driving Reward} (SDR) with a focal-style weighting to amplify reward's value span for improving sensitivity to driving quality. On the {Navsim} benchmark, Curious-VLA achieves state-of-the-art results (PDMS 90.3, EPDMS 85.4) and a Best-of-N PDMS of 94.8, demonstrating its effectiveness in unlocking the exploratory potential of VLA models. Code: \url{https://github.com/Mashiroln/curious_vla.git}.
\end{abstract}
  
\section{Introduction}

\begin{figure}[t]
\centering
    \begin{subfigure}{\linewidth}
      \includegraphics[width=\linewidth]{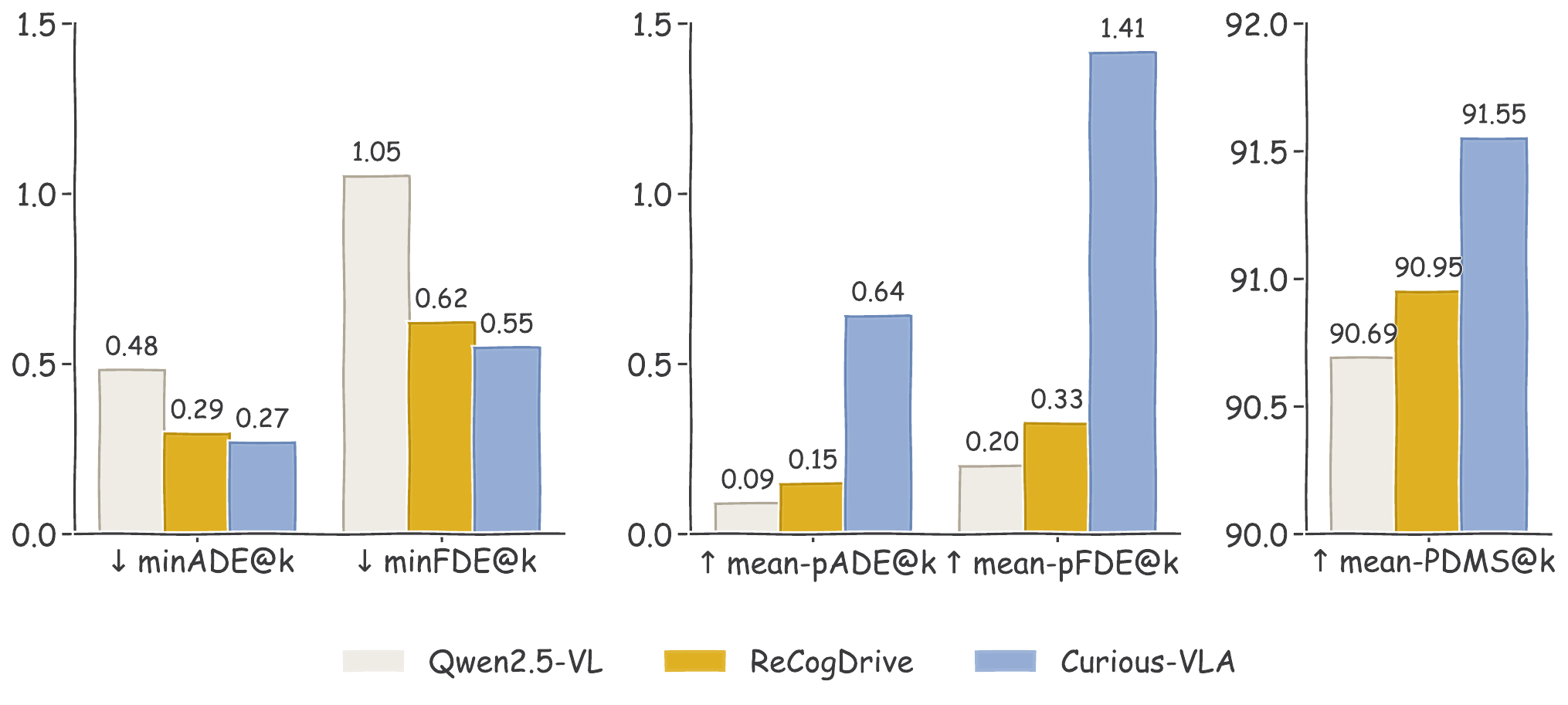}
      \caption{Quantitative comparison for Behavioral Diagnostics.}
      \label{fig:sample_k_stats}
    \end{subfigure}
    \vspace{-6pt}
    \begin{subfigure}{\linewidth}
      \includegraphics[width=\linewidth]{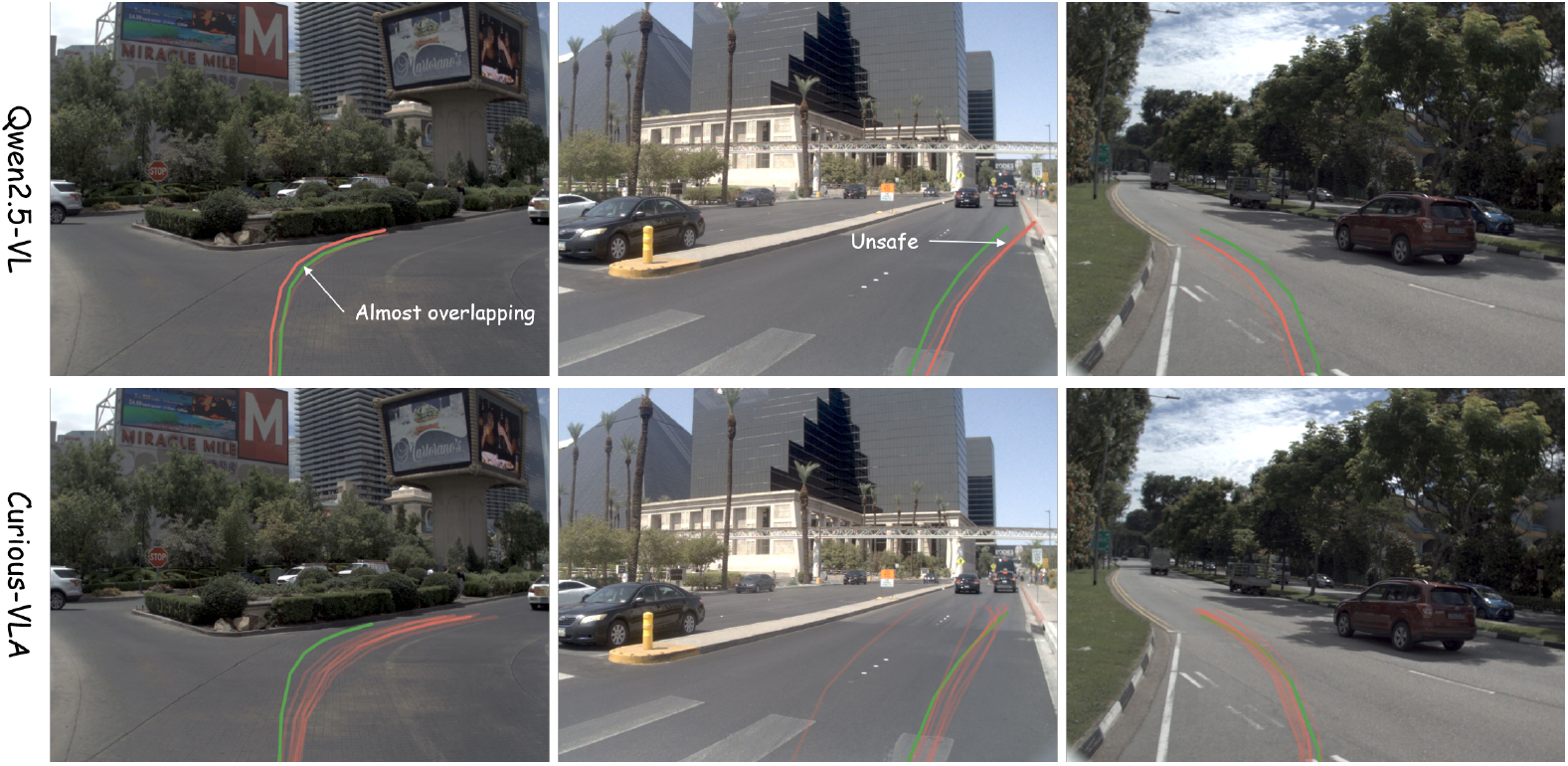}
      \caption{Visualization of multi-infer trajectories, with ($k=8$ in all subfigures)}
      \label{fig:sample_k_vis}
    \end{subfigure}
    \caption{Visualization of Narrow Policy. (a) shows quantitative comparison of exploration. It is obvious that baselines achieve low diversity (mean-pFDE: 0.20m for Qwen2.5-VL, 0.33m for ReCogDrive) and low quality (min-FDE: 1.05m for Qwen2.5-VL). In contrast, our Curious-VLA achieves best diversity, quality and performance. (b) compares the trajectories of Qwen2.5-VL and our Curious-VLA. Qwen2.5-VL trajectories perform exploration collapse with single mode or unsafe behaviors, while ours not.}
    \label{fig:teaser}
\end{figure}
How to effectively leverage Vision-Language Models (VLMs) remains a central research question in autonomous driving.
Recent studies have progressively extended VLMs from perception and understanding to decision-making, facilitating end-to-end {Vision-Language-Action (VLA)} systems for driving. 
Early works such as DriveGPT~\cite{huang2024drivegpt} and LINGO~\cite{Wayve2023LINGO1, marcu2024lingoqa} focused on enhancing driving-scene comprehension with language instruction/behavior. 
Subsequent research has advanced toward reasoning and control, incorporating a richer environmental context~\cite{zhou2025opendrivevla, fu2025orion, li2025recogdrive, li2025drivevlaw0, liu2025occvla} within a single framework.

Current driving VLAs can be broadly categorized into two dominant paradigms:
({i}) {VLA-Planner}~\cite{li2025recogdrive, li2025imagidrive} which relies on an additional trajectory planner module to predict future motion distributions, and ({ii}) {VLA-Token}~\cite{zhou2025autovla, zhou2025opendrivevla, luo2025adathinkdrive} which directly produces trajectory tokens from the LLM decoder. 
Despite architectural differences, both paradigms adhere to a similar two-stage training pipeline: an initial {Imitation Learning (IL)} stage via Supervised Fine Tuning (SFT) to acquire basic trajectory planning and reasoning capability, and a {Reinforcement Learning (RL)} stage with chain-of-thought (CoT) optimization to enhance reasoning~\cite{zhou2025autovla, luo2025adathinkdrive, li2025recogdrive}. 
However, this two-stage pipeline (first IL, then RL) suffers from a fundamental \textbf{Narrow Policy (NP)} limitation, characterized by an inherent exploit–explore imbalance --- {\textit{the IL stage over-exploits ground-truth trajectories, leading to collapsed exploration and consequently restricting policy updates during RL fine-tuning}}. 

This NP problem has been largely neglected in previous driving VLA works. We evaluate two representative baselines, QwenVL-2.5 (VLA-Token) and ReCogDrive (VLA-Planner), on the Navsim \texttt{navtrain} subset~\cite{dauner2024navsim}. 
For each model, we sample $k$ trajectories by $k$ times of inferences, and evaluate them using three \textit{Behavioral Diagnostics} : ({i}) {Diversity}, measured by mean pairwise ADE/FDE, which quantifies trajectory spread; ({ii}) {Quality}, measured by min-ADE/FDE, indicating the best feasible trajectory; ({iii}) {Performance}, equal to mean PDMS of Navsimv1~\cite{dauner2024navsim}. 
As shown in Fig.~\ref{fig:sample_k_stats}, both baselines exhibit evident exploration collapse, with extremely low trajectory diversity and limited trajectory quality. 
Fig.~\ref{fig:sample_k_vis} further illustrates this effect—despite multiple feasible routes, the sampled trajectories converge to a single mode, even leading to unsafe behaviors. 
The narrow policy learned through SFT results in a low-entropy initialization for the subsequent RL stage.  We theoretically analyze this phenomenon in Sec.~\ref{subsec:derivation}.  Since critic-free RL algorithms (\eg, GRPO~\cite{shao2024deepseekmath, guo2025deepseek}) rely on diverse samples to estimate policy gradients, such narrow policy lead to early saturation and limited learning feedback~\cite{yu2025dapo, cui2025entropy}. Consequently, the GRPO RL-training undermines VLA's performances (see Sec 5.5 experiments).

To breakup the limitation, we propose \textbf{Curious-VLA}, a novel training framework that systematically unleashes exploration for VLM itself without any additional module. In the IL stage, we consider the ground-truth (GT) trajectory as just one of the potential human driving behaviors. 
Therefore, we introduce Feasible Trajectory Expansion (FTE) data synthesizing scheme by generating multiple physically valid driving paths, so called feasible trajectory, with VLA-Planner’s diffusion module~\cite{li2025recogdrive}. This data synthesizing scheme largely increases the training trajectory's diversity. 
To accommodate these more diverse trajectories, we normalize each trajectory in a step-wise manner, enhancing the separability of diverse driving behaviors and consequently alleviating the NP problem. 
In the RL stage, to encourage exploration, we further introduce two complementary components: an {Adaptive Diversity-Aware Sampling (ADAS)} strategy and 
the Spanning Driving Reward (SDR).
ADAS prioritizes samples that exhibit exploratory variance by dropping the training examples whose predicted trajectories remain highly similar across multiple inference passes. This encourages the {policy} to refine diverse driving behaviors and prevents premature convergence toward a single dominant pattern. Besides, to further promote effective exploration, we introduce the SDR that reformulates the original driving reward by amplifying  its reward value span through a focal-loss style function, which improves the reward function's sensitivity to driving quality. Our contributions are as follows:

\begin{itemize}
    \item We identify the ``Narrow Policy'' problem, a fundamental bottleneck in the IL-RL pipeline that hinders autonomous driving VLAs. Furthermore, we introduce Behavioral Diagnostics to quantitatively verify this phenomenon.
    \item We propose Curious-VLA, a novel framework that systemically fosters the exploration of VLA model. 
    \item On the Navsim benchmark, Curious-VLA achieves the State-of-The-Art (SoTA) performance of 90.3 PDMS. Furthermore, its Best-of-N PDMS of 94.8 effectively validates that our methods successfully unleash the exploration potential of VLA models.
\end{itemize}

\section{Related Work}

\begin{figure*}[t]
    \centering
    \includegraphics[width=\linewidth]{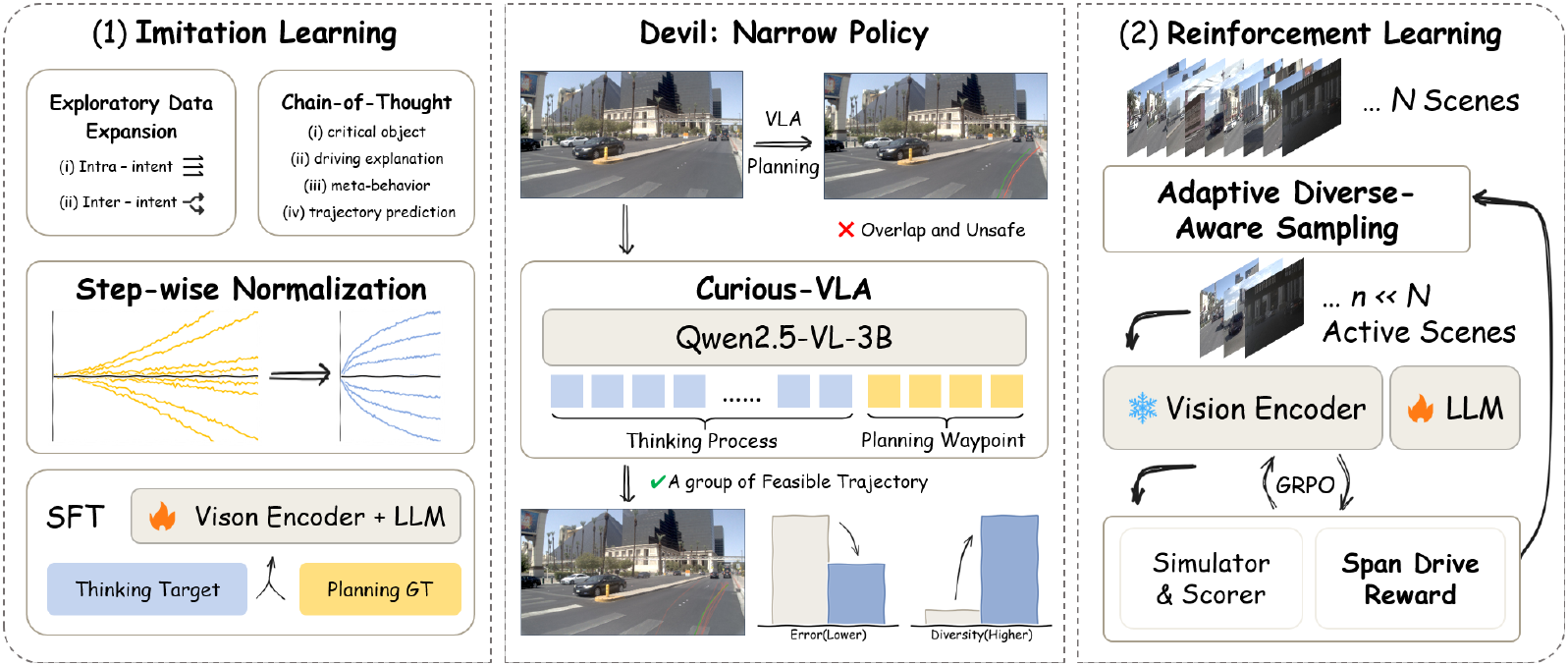}
    \vspace{-15pt}
    \caption{Overall Pipeline of Curious-VLA. {As identified in Sec.~\ref{subsec:derivation}, existing VLAs with the IL-RL pipeline suffer from ``Narrow Policy'' and tend to generate overlap and unsafe behaviors. In this case, we introduce Curious-VLA, which takes into account both the think process and planing waypoint, to improve the diversity, quality and performance (middle panel). Specifically, Curious-VLA alleviates the Narrow Policy in both the IL stage (left panel) and the RL stage (right panel). For IL, to improve the separability of trajectory patterns, we generate diverse trajectories, structure the driving reasoning process into a four-stage CoT, and normalize each prediction step (See Sec.~\ref{subsec:fte}). For RL, to address the advantage collapse problem and sustain exploration, we introduce adaptive diverse aware sampling and spanning driving reward (See Sec.~\ref{subsec:darl}).}}
    \label{fig:main_pipeline}
\end{figure*}

\noindent\textbf{VLA Models for Autonomous Driving.} End-to-End autonomous driving systems~\cite{hu2023planning,jiang2023vad, chitta2022transfuser,li2024hydra,liao2025diffusiondrive,jia2025drivetransformer} are rapidly moving from modular pipelines that decouple driving tasks to perception, prediction, planning to a unified architecture. 
Early research primarily utilized VLMs for {scene understanding and reasoning}, such as captioning, question answering, or intention recognition in driving scenarios~\cite{marcu2024lingoqa, huang2024drivegpt}. 
Recent advancements have extended VLMs to direct action planning, diverging into two paradigms, \textit{VLA-Planner} and \textit{VLA-Token}.
The first paradigm, {{VLA-Planner}} employs the VLM as a semantic reasoner that guides an external planner (\eg, CarLLaVA~\cite{renz2024carllava}, RegCogDrive~\cite{li2025recogdrive}, ORION~\cite{fu2025orion}, ImagiDrive~\cite{li2025imagidrive}, DriveVLA-W0~\cite{li2025drivevlaw0}).
Another paradigm, {VLA-Token} treats trajectory planning as a sequence generation task, where the VLM directly predicts action or waypoint tokens via auto-regressive or diffusion language model. 
Representative approaches based on \textit{text waypoint} include EMMA~\cite{hwang2024emma}, OpenEMMA~\cite{xing2025openemma}, AdaThinkDrive~\cite{luo2025adathinkdrive}, Impromptu-VLA~\cite{chi2025impromptu} and Poutine~\cite{rowe2025poutine}.
In contrast, some methods represent trajectories as \textit{action tokens}, including AutoVLA~\cite{zhou2025autovla} and SMART~\cite{wu2024smart}. 
Despite different paradigms, existing methods suffer from Narrow Policy, primarily stemming from the explore-exploit dilemma, especially the lack of diversity. Our Curious-VLA follows VLA-Token and proposes a systematic framework regarding data, sampling and reward during both IL and RL, thus unlocking the potential of VLA's exploration.

\noindent\textbf{Reinforcement Learning with Verifiable Reward.}
Recent works extend Reinforcement Learning from Human Feedback (RLHF)~\cite{christiano2017deep, ouyang2022training} to {Verifiable Reward} (RLVR)~\cite{shao2024deepseekmath}, optimizing policies by measurable outcomes instead of subjective preferences.
DeepSeek-R1~\cite{guo2025deepseek} and GRPO~\cite{shao2024deepseekmath} eliminate value networks through group-based normalization, improving stability over PPO~\cite{schulman2017ppo}.
Follow-up methods such as DAPO~\cite{yu2025dapo} and KL-CoV/CLIP-Cov~\cite{cui2025entropy} refine advantage estimation or enhance exploration via diversity or curiosity~\cite{dong2025agentic, dai2025cde}.
In autonomous driving, RLVR-style fine-tuning has been adopted in TrajHF~\cite{li2025finetuning}, EvaDrive~\cite{jiao2025evadrive}, AutoVLA~\cite{zhou2025autovla}, and ReCogDrive~\cite{li2025recogdrive}, forming a standard imitation–reinforcement pipeline for VLA models. However, the exploration of RL in VLA is still limited. It is crucial to improve RL's reward mechanisms to enhance exploration. To address this, we form a Diversity-Aware Reinforcement Learning approach by improving both the training data sampling and reward function design.

\section{Preliminary and Narrow Policy}
We first formulate the VLA training pipeline in Sec.~\ref{subsec:pre} and 
then provide an analysis of the neglected Narrow Policy in Sec.~\ref{subsec:derivation}.

\subsection{Preliminary: VLA Training Pipeline}
\label{subsec:pre}

Following prior works~\cite{zhou2025autovla, hu2023planning}, we formulate a driving VLA as a unified generative policy $\pi_\theta$ that maps multimodal observations $\mathcal{X}$ into an action sequence $\tau = \{w_1, \dots, w_T\}$ with length $T$, where $w_i$ represents the ego's spatial and speed action.
Specifically, the multimodal input $\mathcal{X}$ includes multi-view camera images $\mathcal{C}$, textual instructions $\mathcal{I}$ (\eg, ``turn left''), and ego-vehicle states $\mathcal{S}$ (\eg, speed, acceleration, past control actions).  The policy output $\tau$ could be a sequence of waypoints for VLA-Planner or discretized tokens for VLA-Token.
Here, we adopt the VLA-Token paradigm as default, consisting of two stages: Imitation Learning (IL) and Reinforcement Learning (RL).

\noindent\textbf{Imitation Learning.}
In VLA-Token paradigm, the policy is initialized by SFT that maximize the likelihood of generated text output trajectory tokens $\mathbf{y}^*$ via a cross entropy loss:
\begin{equation}
    \label{eq:sft}
    \mathcal{L}_\text{SFT}(\theta) = -\mathop{\mathbb{E}}_{(\mathcal{X}, \mathbf{y}^*) \sim \mathcal{D}} \left[\frac{1}{L} \sum_{t=1}^L \log \pi_\theta(y_t^* | \mathbf{y}^*_{<t}, \mathcal{X}) \right],
\end{equation}
which establishes VLA's basic planning capabilities.

\noindent\textbf{Reinforcement Learning.}
To foster genuine environmental understanding and active driving reasoning,
the SFT policy $\pi_\text{sft}$ is further refined using RL.
For RL, Group Relative Policy Optimization (GRPO)~\cite{shao2024deepseekmath} is commonly used and can eliminate the value network by normalizing the advantages within a sampled group.
For each input $\mathcal{X}$, a group of $G$ outputs $\{\mathbf{y}_i\}_{i=1}^G$
is sampled from the old policy $\pi_{\theta_\text{old}}$.
The training objective combines a clipped loss and a KL divergence constraint:
\begin{equation}
\begin{aligned}
    \label{eq:grpo}
    &\mathcal{J}_\text{GRPO}(\theta) = \\
    &~~~~\mathop{\mathbb{E}}_{\mathcal{X}, \{\mathbf{y}_i\} \sim \pi_{\theta_\text{old}}} \bigg[ \frac{1}{G} \sum_{i=1}^G \Big( \hat{\mathcal{J}}_i(\theta) 
    - \beta \mathbb{D}_\text{KL}(\pi_\theta || \pi_\text{sft}) \Big) \bigg],
\end{aligned}
\end{equation}
where $\hat{\mathcal{J}}_i(\theta)$ is the clipped advantage term:
\begin{equation}
    \label{eq:pg_loss_a}
    \hat{\mathcal{J}}_i(\theta) = \min \left( \rho_i(\theta) A_i, \text{clip}\left(\rho_i(\theta), 1-\epsilon, 1+\epsilon\right) A_i \right),
\end{equation}
with likelihood ratio $\rho_i(\theta) = \frac{\pi_\theta(\mathbf{y}_i|\mathcal{X})}{\pi_{\theta_\text{old}}(\mathbf{y}_i|\mathcal{X})}$.
The advantage $A_i$ is computed by standardizing the group rewards:
\begin{equation}
    \label{eq:grpo_adv_func}
    A_i = \frac{R(\mathbf{y}_i) - \mu_R}{\sigma_R + \xi}.
\end{equation}

Here, $\mu_R, \sigma_R$ are the mean and standard deviation of rewards within the group.

\subsection{Analysis of the Narrow Policy (NP)}
\label{subsec:derivation}
We first analyze the emergence of NP problems in (1)-(3). Then, we provide practical analysis metrics in (4) towards the NP problem. 

\noindent\textbf{(1) Optimization Objective Mismatch.}
The cross-entropy loss in Eq.~\ref{eq:sft} treats all non-ground-truth tokens as equally incorrect~\cite{li2022generalized}—%
it lacks any notion of spatial or functional proximity between trajectory tokens, which should be physically continuous. Formally, let $z_{t}$ be the model's output logit, $\hat{y}_t$ be the predicted token and $y_{t}^{*}$ be the ground-truth (GT) token at step $t$. $\hat{y}_t$ is generated using the causal context $\{\mathbf{y}^*_{<t}, \mathcal{X}\}$. The gradient of the total loss $\mathcal{L}_\text{SFT}$ with respect to this single logit $z_{t}$ is:
\begin{equation}
\frac{\partial \mathcal{L}_\text{SFT}}{\partial z_{t}} 
= \pi_\theta(y_k | \mathbf{y}^*_{<t}, \mathcal{X}) - \mathbb{I}(\hat{y}_t = y_t^*).
\end{equation}

While the penalty magnitude scales with the model’s confidence $\pi_\theta(y_t|\cdot)$,
{the optimization objective itself offers no smoother incentive for near-correct predictions over clearly erroneous ones.} For example, compared with regression loss, when the ground truth is \texttt{31.5}, it is unclear which CE loss would be higher for the discrete tokens \texttt{31.4} and \texttt{21.4}.
This discrete, per-token supervision encourages overconfidence in $\mathbf{y}^*$,
collapsing the policy distribution around a single expert mode.

\noindent\textbf{(2) Horizon Physical Scale Mismatch during SFT.} The other bottleneck is rooted in the trajectory representation itself. Fig.~\ref{fig:divergence} shows that future waypoints are predicted in an ego-centric coordinate frame, where distant horizons exhibit larger spatial variance. For example, the variance of the waypoints' coordinates at $t=4s$ are orders of magnitude larger than that at $t=0.5s$. In result, far-horizon losses dominate $\mathcal{L}_\text{SFT}$, while near-horizon actions (which determine steering precision) contribute negligibly. This imbalance reduces VLA's capability to learn behavioral diversity.

\noindent\textbf{(3) Advantage Collapse in RL.}
When the policy $\pi_\theta$ collapses to a single trajectory mode, rewards become nearly identical across samples, \ie, $R(\mathbf{y}_i) \approx \mu_R$.  
Consequently, $\sigma_R \to 0$ and thus $A_i \approx 0$:
\begin{equation}
\lim_{\sigma_R \to 0} A_i =
\frac{R(\mathbf{y}_i)-\mu_R}{\sigma_R+\xi} \to 0,
\end{equation}
leading to vanishing gradients in Eq.~\ref{eq:grpo}.

\noindent\textbf{(4) Behavioral Diagnostics.}
To quantitatively diagnose the {narrow policy} phenomenon, we introduce three complementary metrics collectively referred to as {Behavioral Diagnostics}.  
Given an input scenario $\mathcal{X}$, we sample $k=8$ trajectories from the policy $\pi_\theta$, producing a trajectory set 
$\mathcal{T} = \{\tau_1, \tau_2, \dots, \tau_k\}$ over a 4-second (8-step) horizon.  
Let $\tau^*$ denote the ground-truth trajectory.  
The diagnostics are defined as follows:

\begin{itemize}
\item {Diversity}: Measures the spread of the policy’s exploration. 
We compute the mean pairwise Average Displacement Error (pADE) or Final Displacement Error (pFDE) between all sampled trajectories in $\mathcal{T}$.  
A lower value indicates limited behavioral diversity and thus reduced exploration capacity.
\item {Quality}: 
Evaluates the best feasible outcome within the sampled set.  
It is measured by the minimum ADE/FDE with respect to $\tau^*$ across all trajectories in $\mathcal{T}$.  This metric reflects whether diverse exploration still preserves optimal planning quality.
\item {Performance}: 
Assesses overall driving competence using the mean PDMS~\cite{dauner2024navsim} score from the Navsimv1 benchmark, which integrates safety, comfort, and efficiency.
\end{itemize}

Together, these diagnostics reveal both the {breadth} and {effectiveness} of policy exploration. A well-balanced model should exhibit high Diversity@$k$, low Quality@$k$ (indicating at least one good sample), and high Performance@$k$.  
In contrast, the collapse of Diversity@$k$ alongside stagnant Quality@$k$ directly signals the onset of the NP bottleneck. The experimental results are in {Sec.~\ref{subsec:analysis_diversity}}.

\section{Curious-VLA}
The overall pipeline of Curious-VLA is shown in Fig.~\ref{fig:main_pipeline}. Specifically, Curious-VLA consists of Feasible Trajectory Expansion in Imitation Learning and Diversity-Aware Reinforcement Learning.

\subsection{Feasible Trajectory Expansion in Imitation Learning}
\label{subsec:fte}
To address the narrow policy problem rooted in IL, we design Feasible Trajectory Expansion (FTE) to balance the explore–exploit trade-off. FTE builds on standard SFT and comprises 1) Exploratory Data Expansion (DE), 2) Chain-of-Thought Data Synthesis (CoT), and 3) Step-wise Normalization (SN).

\noindent\textbf{Exploratory Data Expansion.}
\label{subsec:data}
We first identify 12k challenging driving segments (multi-lane, intersection, occlusion) from the 103k NavTrain set~\cite{dauner2024navsim} using Qwen2.5-VL-72B filtering.  
Then, leveraging diffusion-based ReCogDrive~\cite{li2025recogdrive}, we generate diverse trajectories by perturbing diffusion latents.  
All candidate trajectories are filtered using the PDMS scorer to ensure safety compliance.  
FTE expands data both \textit{within-intent} (sampling around the same driving goal) and \textit{across-intent} (altering route-level decisions), resulting in 142k safe and diverse samples.

\noindent\textbf{Chain-of-Thought Data Synthesis.} 
Following previous methodology~\cite{sima2024drivelm, rowe2025poutine},
we structure the driving reasoning process into a four-stage chain in single-turn dialogue:
(i) critical object perception, (ii) driving explanation,
(iii) meta-behavior description, and final (iv) trajectory prediction.
We leverage Qwen2.5-VL-72B to automatically generate these structured reasoning sequences
for our entire expanded dataset. The implementation details are available in the Supplement.

\begin{figure}[t]
\centering
\includegraphics[width=\linewidth]{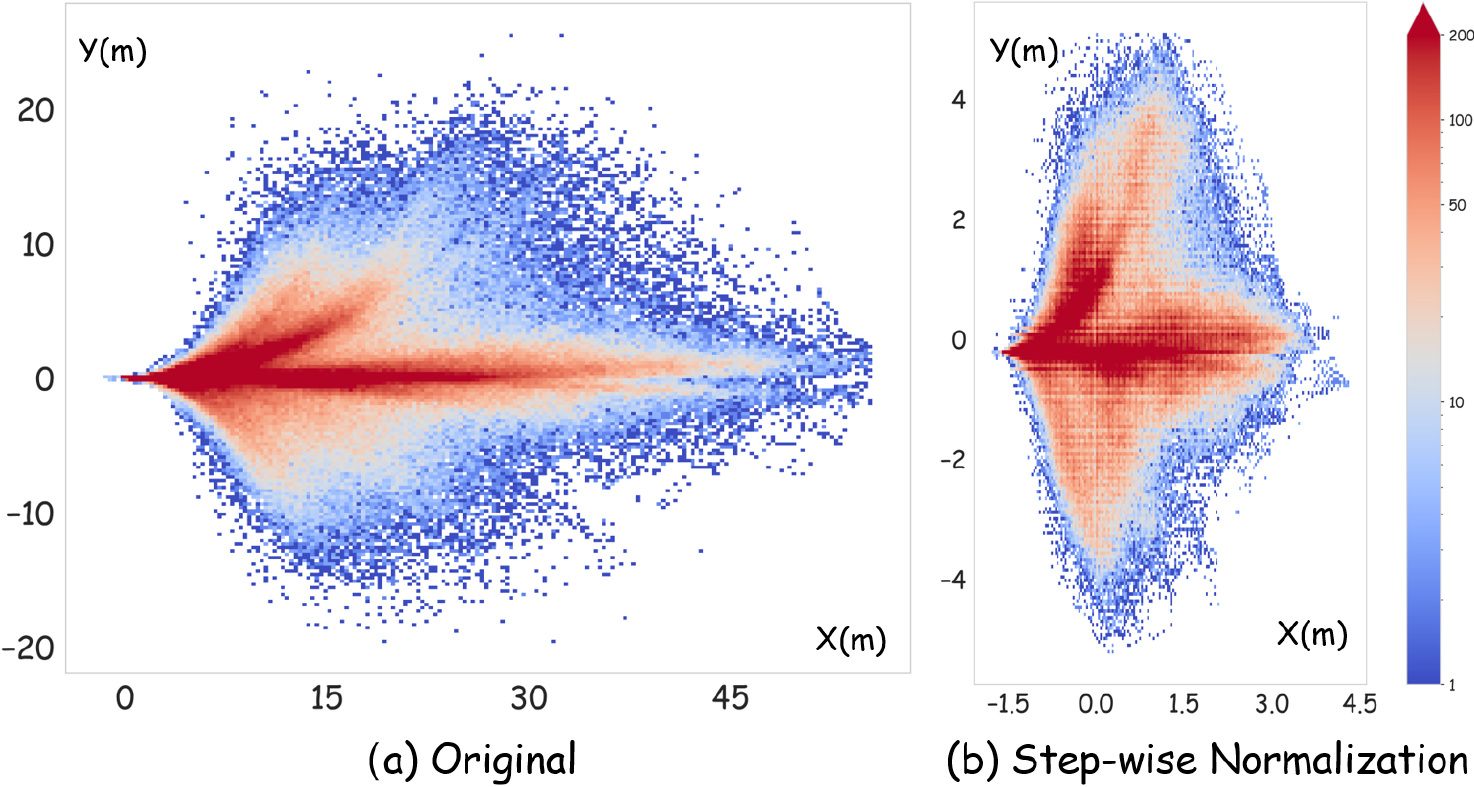}
\caption{Visualization of horizon physical scale mismatch by waypoints distribution.}
\label{fig:divergence}
\end{figure}
\noindent\textbf{Step-wise Normalization.}
To handle the horizon-scale imbalance mentioned in Sec.~\ref{subsec:derivation}~(2), we normalize each prediction step $t$ independently:
\begin{equation}
\tilde{w}_t = \frac{w_t - \mu_t}{\sigma_t}, \quad
\hat{w}_t = \hat{\tilde{w}}_t \sigma_t + \mu_t,
\end{equation}
where $(\mu_t,\sigma_t)$ are per-step statistics from the training set. 
During SFT, we use $\tilde{w}_t$ for training. For testing, the predicted $\hat{\tilde{w_t}}$ need to de-normalize to trajectory $\hat{w_t}$.
Fig.~\ref{fig:divergence} shows that this normalization equalizes gradient magnitudes across horizons, improving the separability of trajectory patterns and providing a balanced foundation for exploration.

\subsection{Diversity-Aware Reinforcement Learning}
\label{subsec:darl}

To sustain exploration in RL, we introduce two complementary mechanisms:
{Adaptive Diversity-Aware Sampling (ADAS)} and {Spanning Driving Reward (SDR)}.

\begin{algorithm}[t]
\caption{Adaptive Diversity-Aware Sampling (ADAS)}
\label{alg:adas}
\SetAlgoLined
\KwIn{Full dataset $\mathcal{D}_{total}$, initial policy $\pi_0$}
\textbf{Hparam:} Offline rollout size $M$, group size $G$\, Diversity threshold $\epsilon_{div}$, Confidence margin $\epsilon_{conf}$\;
\KwOut{Refined policy $\pi_{E}$}
\For{outer-loop $e = 1, \dots, E$}{
    \tcp{Phase 1: Offline Filtration}
    Initialize active set $\mathcal{D}_{active} \leftarrow \emptyset$\;
    \For{scenario $x \in \mathcal{D}_{total}$}{
        Generate $M$ offline rollouts using $\pi_{e-1}$\;
        Estimate reward stats ($\mu_R, \sigma_R$) and success rate $p \leftarrow \mu_R / R_{max}$\;
        \If{$p^G + (1-p)^G < \epsilon_{div}$ \textbf{and} $|\sigma_R - \sigma_{Bernoulli}| < \epsilon_{conf}$}{
            $\mathcal{D}_{active} \leftarrow \mathcal{D}_{active} \cup \{x\}$\;
        }
    }
    \tcp{Phase 2: GRPO Training}
    \While{not end of epoch}{
        Sample batch of scenarios $\mathcal{B} \subset \mathcal{D}_{active}$\;
        For each $x \in \mathcal{B}$, generate online group of size $G$ using $\pi_{e-1}$\;
        Update $\pi_e$ via policy gradient (Eq.~\ref{eq:grpo})\;
    }
}
\Return{$\pi_E$}
\end{algorithm}

\noindent\textbf{Adaptive Diversity-Aware Sampling.}
ADAS dynamically \textit{selects scenarios that yield diverse rollouts under stochastic policies, maintaining sufficient reward variance for stable GRPO optimization} as discussed in Sec.~\ref{subsec:derivation} (3). 

We model the outcome variability of each scenario as a simplified Bernoulli process, where each rollout corresponds to a binary trial of \textit{success} (high PDMS) or \textit{failure} (low PDMS) with probability $p$.  
This approximation captures the most extreme case of reward distribution, providing a simple yet effective measure of diversity potential.
At the beginning of each training outer-loop, we re-sample a new active training set from entire train data.
For each training scenario $x$, we periodically perform $M$ offline rollouts ($M \gg G$) using the current policy to estimate its empirical reward distribution.  
The average normalized PDMS across these rollouts serves as the success probability estimate $\hat{p}$. A scenario $x$ is included in the active training set only if it satisfies two diversity-related conditions:
\begin{equation}
\begin{aligned}
&\hat{p}^G + (1-\hat{p})^G < \epsilon_\text{div}, 
\\
&|\sigma_R - \sqrt{\hat{p}(1-\hat{p})}R_\text{range}| < \epsilon_\text{conf},
\label{eq:adas_filter}
\end{aligned}
\end{equation}
where $\epsilon_\text{div}$, $\epsilon_\text{conf}$ are predefined thresholds. The first term bounds the probability that all $G$ online rollouts yield identical outcomes (either all success or all failure), ensuring sufficient variability across samples.  The second term enforces consistency between the empirical standard deviation $\sigma_R$ and the theoretical Bernoulli variance $\sqrt{p(1-p)}R_\text{range}$ within a confidence margin $\epsilon_\text{conf}$, filtering out unstable or noisy scenarios.

\noindent\textbf{Spanning Driving  Reward.}
\label{subsec:sdr}
To further amplify the exploration signals, we redesign the reward based on the Navsim metrics PDMS and EPDMS~\cite{dauner2024navsim}.  Each metric is computed as the product of safety constraints ($C$) and weighted objectives ($M$):
\begin{equation} 
\text{PDMS} = \prod_{c \in C} c \times \frac{\sum_{m \in M} w_m \cdot m}{\sum_{m \in M} w_m}, 
\end{equation}
where $C = \{\text{NC, DAC}\}$ (No Collisions, Drivable Area Compliance) and $M = \{\text{EP, TTC, C}\}$ (Ego Progress, Time to Collision, Comfort) with weights $w_m = \{5, 5, 2\}$. We reformulate this into a {focal-style spanning objective}:
\begin{equation}
\label{eq:r_sparse} R_\text{span} = \prod_{c \in C} c \cdot 
\frac{\sum_{m \in M} w'_m \cdot (1 - (1 - m)^{\gamma_m})}{\sum_{m \in M} w'_m},
\end{equation} where $\gamma_{m}$ is the hyperparameter. 
The EPDMS~\cite{dauner2024navsim, cao2025pseudo} reuses the same calculation structure and extends $C$ with $\{\text{DDC, TLC}\}$ (Driving Direction, Traffic Light Compliance), extends $M$ with $\{\text{LK, EC}\}$ (Lane Keeping, two-frame Extended Comfort), with the extra weights $\{2, 2\}$. This focal-style design magnifies differences between suboptimal and optimal behaviors, improving the reward's sensitivity to the driving quality.

\section{Experiment}

\begin{figure*}[t]
    \centering
    \includegraphics[width=0.9\linewidth]{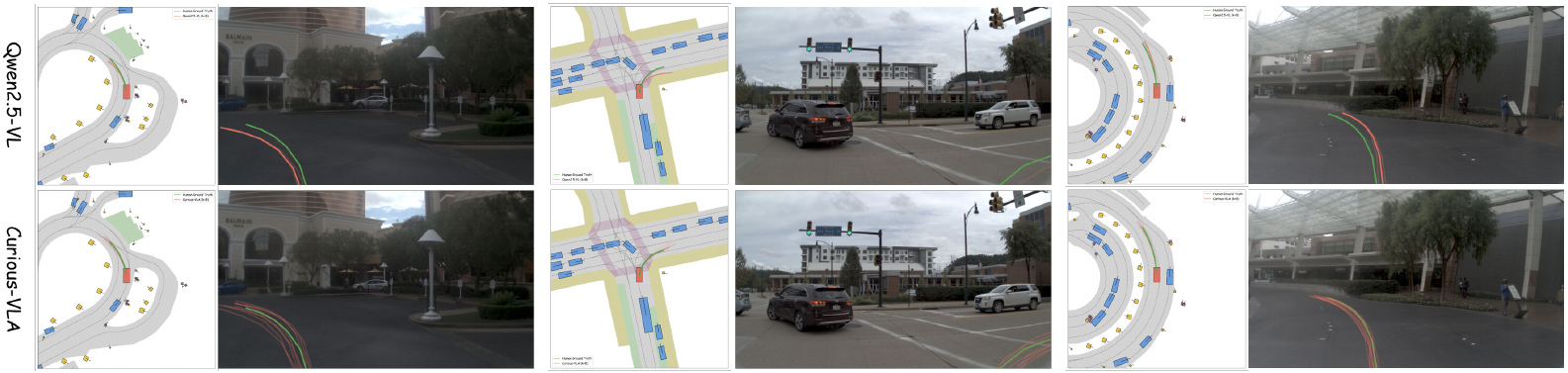}
    \caption{Qualitative comparison with BEV and Camera. We can see that our Curious-VLA achieves more feasible trajectories.}
    \label{fig:more_vis}
\end{figure*}

\begin{table*}[ht]
\centering
\caption{Comparison of different methods on Navsim V1 benchmark. The model with $^{\dagger}$ indicates evaluation using Best-of-N sampling. The best and second best results are \textbf{bolded} and \underline{underlined}, respectively. \texttt{Continuous} regresses the trajectory with an additional MLP/Transformer head. \texttt{Discrete Action} decodes action tokens that discretize each dimension of the trajectory into several bins. \texttt{Text Waypoint} decodes textual numbers that represent the trajectory.}
\label{tab:main_pdms}
\resizebox{0.85\textwidth}{!}{%
\begin{tabular}{lccccccccccc}
\toprule
& Base Model & Sensors & Trajectory& NC$\uparrow$ & DAC$\uparrow$ & EP$\uparrow$ & TTC$\uparrow$ & C$\uparrow$ & PDMS$\uparrow$ \\
\midrule
\multicolumn{9}{l}{\textbf{\textit{Human GT}}} \\
\textcolor{gray}{Human}~\cite{dauner2024navsim} & - & - & -& \textcolor{gray}{100.0} & \textcolor{gray}{100.0} &
\textcolor{gray}{87.5} & \textcolor{gray}{100.0} &
\textcolor{gray}{99.9} & \textcolor{gray}{94.8} \\
\midrule
\multicolumn{9}{l}{\textbf{\textit{Classical E2E}}} \\
{Ego-MLP~\cite{li2024ego}} & - & 6x C & Continuous & 93.1 & 78.3 & 63.2 & 84.0 & 100.0 & 66.4 \\
{UniAD~\cite{hu2023planning}} & - & 3x C + L & Continuous & 97.7 & 92.8 & 79.2 & 92.8 & 100.0 & 84.0 \\
{DiffusionDrive~\cite{liao2025diffusiondrive}} & - & 3x C + L & Continuous & 98.2 & 96.2 & 82.2 & 94.7 & 100.0 & 88.1 \\
{WoTE~\cite{li2025end}} & - & 3x C + L & Continuous  & 98.5 & 96.8 & 81.9 & 94.4 & 99.9 & 88.3 \\
\midrule
\multicolumn{9}{l}{\textbf{\textit{VLA(VLA-Planner)}}} \\
ImagiDrive-A~\cite{li2025imagidrive} & InternVL2.5-4B & 1x C & Continuous  & 98.1 & 96.2 & 80.1 & 94.4 & \textbf{100.0} & 86.9 \\
ImagiDrive-S & InternVL2.5-4B & 1x C & Continuous & 98.6 & 96.2 & 80.5 & 94.5 & \textbf{100.0} & 87.4 \\
ReCogDrive~\cite{li2025recogdrive} & InternVL2-8B & 3x C & Continuous & 98.2 & 97.8 & 83.5 & 95.2 & \textbf{100.0} & 89.6 \\
DriveVLA-W0~\cite{li2025drivevlaw0} & Emu-3-8B & 1x C & Continuous & 98.7 & \textbf{99.1} & \textbf{93.3} & 95.3 & 99.3 & 90.2 \\
DriveVLA-W0$^{\dagger}$ & Emu-3-8B & 1x C & Continuous & \underline{99.3} & 97.4 & 88.3 & 97.0 & \underline{99.9} & \underline{93.0} \\
\midrule
\multicolumn{9}{l}{\textbf{\textit{VLA(VLA-Token)}}} \\
Qwen2.5-VL~\cite{li2025recogdrive} & Qwen2.5-VL-7B & 1x C & Text Waypoint  & 97.8 & 92.1 & 78.3 & 92.8 & \textbf{100.0} & 83.3 \\
InternVL2~\cite{li2025recogdrive} & InternVL2-8B & 1x C & Text Waypoint & 97.0 & 92.4 & 78.9 & 91.8 & \textbf{100.0} & 83.3 \\
AutoVLA~\cite{zhou2025autovla} & Qwen2.5-VL-3B & 3x C & Discrete Action & 98.4 & 95.6 & 81.9 & \underline{98.0} & \underline{99.9} & 89.1 \\
AutoVLA$^{\dagger}$ & Qwen2.5-VL-3B & 3x C & Discrete Action & 99.1 & 98.8 & 87.9 & 97.2 & \textbf{100.0} & 92.1 \\
AdaThinkDrive~\cite{luo2025adathinkdrive} & InternVL3-8B & 1x C & TextWaypoint & 98.4 & 97.8 & 84.4 & 95.2 & \textbf{100.0} & 90.3 \\
AdaThinkDrive$^{\dagger}$ & InternVL3-8B & 1x C & Text Waypoint & 99.1 & 98.8 & 87.9 & 95.2 & \textbf{100.0} & \underline{93.0} \\
\midrule
\textbf{Curious-VLA} & Qwen2.5-VL-3B & 1x C & Text Waypoint & 98.4 & 96.9 & 88.5 & 97.9 & 98.1 & 90.3 \\
\textbf{Curious-VLA}$^{\dagger}$ & Qwen2.5-VL-3B & 1x C & Text Waypoint & \textbf{99.5} & \underline{99.0} & \underline{91.8} & \textbf{99.3} & 98.4 & \textbf{94.8} \\
\bottomrule
\end{tabular}}
\end{table*}

\begin{table*}[t]
\centering
\caption{Comparison of different methods with the Extended PDMS metric of NAVISM V2.}
\label{tab:main_epdms}
\scalebox{0.9}{
\begin{tabular}{l|ccccccccc|c}
\toprule
Method & NC $\uparrow$ & DAC $\uparrow$ & DDC $\uparrow$ & TLC $\uparrow$ & EP $\uparrow$ & TTC $\uparrow$ & LK $\uparrow$ & C $\uparrow$ & EC $\uparrow$ & \textbf{EPDMS} $\uparrow$ \\
\midrule
Ego-MLP~\cite{li2024ego} & 93.1 & 77.9 & 92.7 & 99.6 & 86.0 & 91.5 & 89.4 & 98.3 & 85.4 & 64.0 \\
VADv2~\cite{jiang2023vad} & 97.3 & 91.7 & 98.2 & \textbf{99.9} & 77.6 & 92.7 & 66.0 & \textbf{100.0} & \textbf{97.4} & 76.6 \\
TransFuser~\cite{chitta2022transfuser} & 96.9 & 89.9 & 97.8 & 99.7 & 87.1 & 95.4 & 92.7 & 98.3 & 87.2 & 76.7 \\
HydraMDP++~\cite{li2024hydra} & 97.2 & 97.5 & 99.4 & 99.6 & 83.1 & 96.5 & 94.4 & 98.2 & 70.9 & 81.4 \\
DriveSuprim~\cite{yao2025drivesuprim} & 97.5 & 96.5 & 99.4 & 99.6 & 88.4 & 96.6 & 95.5 & 98.3 & 77.0 & 83.1 \\
ARTEMIS~\cite{feng2025artemis} & 98.3 & 95.1 & 98.6 & 99.8 & 81.5 & 97.4 & 96.5 & 98.3 & -- & 83.1 \\
ReCogDrive~\cite{li2025recogdrive} & 98.3 & 95.2 & \textbf{99.5} & 99.8 & 87.1 & 97.5 & 96.6 & 98.3 & 86.5 & 83.6 \\
DiffusionDrive~\cite{liao2025diffusiondrive} & 98.2 & 95.9 & 99.4 & 99.8 & 87.5 & 97.3 & 96.8 & 98.3 & 87.7 & 84.5 \\
\midrule
\textbf{Curious-VLA} & \textbf{98.4} & \textbf{96.9} & 99.2 & 99.8 & \textbf{88.5} & \textbf{97.9} & \textbf{96.9} & 98.1 & 81.5 & \textbf{85.3} \\
\bottomrule
\end{tabular}}
\vspace{-5px}
\end{table*}

\begin{table}[t]
\centering

\setlength{\tabcolsep}{5pt}
\caption{Open-loop evaluation on the nuScenes Benchmark.}
\label{tab:nuscenes_l2}
\resizebox{0.48\textwidth}{!}{
\begin{tabular}{lcccc}
\toprule
\multirow{2}{*}{Method} & \multicolumn{2}{c}{ST-P3 metrics} & \multicolumn{2}{c}{UniAD metrics} \\
\cmidrule(lr){2-3} \cmidrule(lr){4-5}
 & L2 (m) $\downarrow$ & Collision (\%) $\downarrow$ & L2 (m) $\downarrow$ & Collision (\%) $\downarrow$ \\
\midrule
ST-P3~\cite{hu2022st} & 2.11 & 0.71 & -- & -- \\
VAD~\cite{jiang2023vad} & 0.37 & 0.14 & -- & -- \\
UniAD~\cite{hu2023planning} & 0.69 & 0.12 & 1.03 & 0.31 \\
EMMA~\cite{hwang2024emma} & 0.32 & -- & -- & -- \\
OpenEMMA~\cite{xing2025openemma} & 2.81 & -- & -- & -- \\
OpenDriveVLA~\cite{zhou2025opendrivevla}& 0.33 & 0.10 & 0.67 & 0.30 \\
AutoVLA~\cite{zhou2025autovla} & 0.48 & 0.13 & 0.86 & 0.35 \\
Impromptu VLA~\cite{chi2025impromptu} & 0.33 & 0.13 & 0.67 & 0.38\\
\midrule
Curious-VLA  & \textbf{0.31} & \textbf{0.10} & \textbf{0.60} & \textbf{0.33} \\
\bottomrule
\end{tabular}}
\end{table}

\subsection{Datasets and Evaluation Metrics}

\noindent\textbf{Navsim (v1/v2).} We conduct experiments on Navsim-v1~\cite{dauner2024navsim} and v2~\cite{cao2025pseudo}, which are built on OpenScene~\cite{yang2023vidar, openscene2023, sima2023_occnet} (a redistribution of nuPlan~\cite{nuplan}) for large-scale, non-reactive simulation. Agent inputs include 8 surround-view cameras and 5 LiDAR sensors, optionally with up to 3 historical frames (1.5s  @ 2Hz). We build our training set from the only \textit{CAMERA-FRONT} view of the official \texttt{navtrain} split ($\sim$103k samples), following the procedure described in Sec.~\ref{subsec:data}.
For evaluation, we report the official \textbf{PDMS} (for v1) and \textbf{EPDMS} (for v2) scores, which are formally described in Sec.~\ref{subsec:darl}.

\noindent\textbf{NuScenes.} To verify real-world generalization, we conduct supplementary experiments on nuScenes. This dataset contains 1000 20s scenes ($\sim$1.4M camera images and $\sim$390k LiDAR scans). Following the evaluation protocols of UniAD~\cite{hu2023planning} and ST-P3~\cite{hu2022st}, we report L2 distance error (L2) and Collision rate for direct comparison.

\subsection{Implement Details}

Curious-VLA is a pure VLM that directly auto-regresses text waypoint trajectories without any additional planner module, initialized from Qwen2.5-VL-3B~\cite{bai2025qwen2}. Our two-stage training pipeline consists of: 1) An Imitation Learning (by SFT) stage using \texttt{LLaMA-Factory}~\cite{zheng2024llamafactory}, and 2) A Reinforcement Learning stage adapted from \texttt{VeRL}\cite{sheng2024hybridflow} and \texttt{EasyR1}\cite{zheng2025easyr1}.All experiments are conducted on 8 NVIDIA H100 GPUs using DeepSpeed ZeRO-1~\cite{rajbhandari2020zero}. 
For SFT, we train for 6 total epochs with a global batch size of 128. For RL, we train for 130 steps in total with 3 outer-loop for ADAS. The count of rollout is 8 and the actor global batch size is 256. More details can be found in the supplementary materials.

\subsection{Main Results}

\noindent\textbf{Open-loop Results on Navsim v1 Benchmark.}
As shown in Tab.~\ref{tab:main_pdms}, Curious-VLA reaches a PDMS of 90.3, establishing a new SOTA record under the single-front-camera input on Navsim v1.
It not only far exceeds traditional E2E methods using Camera+LiDAR inputs (e.g., WoTE~\cite{li2025end} at 88.3) but also demonstrates remarkable efficiency among VLM-based approaches. Notably, despite using a smaller (3B) and earlier (Qwen2.5-VL) base model, Curious-VLA's performance is comparable to the VLA-Planner method (DriveVLA-W0~\cite{li2025drivevlaw0} at 90.2), and the VLA-Token method (AdaThinkDrive~\cite{luo2025adathinkdrive}, 90.3). Compared to AutoVLA~\cite{zhou2025autovla} which uses the same base model, our method achieves a significant 1.2 PDMS improvement (90.3 vs. 89.1). Although the Comfort (C) metric is slightly reduced, the synchronous enhancement in both Ego Progress (EP) and Time to Collision (TTC) is strong evidence of our success in overcoming key performance obstacles.

More importantly, following the AutoVLA Best-of-N setting (N=6), Curious-VLA$^\dagger$ achieves a PDMS score of 94.8. This result surpasses the AdaThinkDrive$^\dagger$ (93.0) by 1.8 PDMS and matches the Human GT level. This exceptional performance stems from our model's ability to predict trajectories with both high diversity and high quality, enabling it to make diverse yet correct decisions in complex scenarios. It effectively validates that our unleashing exploration mechanism has successfully alleviated the narrow policy problem. More visualization in ~\ref{fig:more_vis}.

\noindent\textbf{Open-loop Results on Navsim v2 Benchmark.} We further evaluated Curious-VLA on the more challenging Navsim v2, using the official \texttt{navtest} split. As shown in Tab.~\ref{tab:main_epdms}, our method achieves an EPDMS composite score of 85.3, establishing a new SOTA. This represents a +0.8 improvement over the DiffusionDrive (84.5). The score is supported by our model's robust performance across several critical sub-metrics, particularly in Drivable Area Compliance (DAC), Ego Progress (EP), and Time to Collision (TTC).

\noindent\textbf{Open-loop Results on Nuscenes Benchmark.} We applied our training pipeline to a 28k Nuscenes training subset~\cite{chi2025impromptu}, using an ADE-based reward ~\cite{zhou2025autovla} for the RL stage. As shown in Tab.~\ref{tab:nuscenes_l2}, Curious-VLA also achieves excellent results on L2 error and Collision rate in 3s, outperforming existing VLAs and E2E models.

\subsection{Analytical Experiments of Exploration.}
\label{subsec:analysis_diversity}

 To analyze how our method progressively addresses the narrow policy problem, we use the Behavioral Diagnostics evaluated on the \texttt{navtrain} subset. In in Tab.~\ref{tab:exploration_diagnostics}. Both the Qwen2.5-VL baseline and ReCogDrive suffer from severely limited diversity. While adding expanded CoT data alone (+ FTE, w/o SN) degrades trajectory quality, introducing Step-wise Normalization (+ FTE) significantly boosts diversity while maintaining high performance (91.31 mean-PDMS). This shows that SN is the key to effectively learning from diverse trajectories. Finally, the RL stage (+ FTE + RL) further pushes both diversity and quality to the best levels (\textbf{1.415} mean-pFDE and \textbf{0.547} minFDE).

\begin{table}[ht]
\footnotesize
\centering
\caption{Exploration analysis (all metrics @k=8). We evaluate \textbf{Quality} (minADE/FDE $\downarrow$), \textbf{Diversity} (mean-pADE/FDE $\uparrow$), and \textbf{Perf.} (mean-PDMS $\uparrow$). In the Quality and Diversity columns, values correspond to ADE / FDE.}
\label{tab:exploration_diagnostics}
\begin{tabular}{llccc}
\toprule
\textbf{Method} & {\textbf{Stage}} & \textbf{Quality} & \textbf{Diversity} & \textbf{Perf.} \\
\midrule
ReCogDrive & IL+RL & 0.295 / 0.621 & 0.148 / 0.325 & 90.95 \\
\midrule
Qwen2.5-VL & IL & 0.481 / 1.052 & 0.090 / 0.200 & 90.69 \\
+ FTE (w/o SN) & IL & 0.513 / 1.129 & 0.170 / 0.381 & 90.65 \\
+ FTE & IL & 0.480 / 1.078 & 0.346 / 0.803 & 91.31 \\
+ FTE + RL & IL+RL & \textbf{0.269 / 0.547} & \textbf{0.641 / 1.415} & \textbf{91.55} \\
\bottomrule
\vspace{-25px}
\end{tabular}%
\end{table}

\subsection{Ablation Study}

\noindent\textbf{Ablation on Feasible Trajectory Expansion.}
We perform an IL-stage ablation to assess the contribution of our Feasible Trajectory Expansion in Tab.~\ref{tab:ablation_il_components}. Following standard RL practice that initializes from high-quality reasoning models (\eg, \cite{ouyang2022training, guo2025deepseek, zhou2025autovla}), we first build a strong baseline by adding CoT supervision. This yields a performance gain of $\mathbf{+1.7}$ PDMS and serves as the basis for subsequent ablations.

FTE comprises two other training designs: Exploratory Data Expansion (DE) and Step-wise Normalization (SN). The core challenge is to leverage DE effectively: adding DE alone (without SN) yields ${85.2}$ PDMS, which is worse than the CoT baseline ($85.6$), indicating that naive data expansion is difficult to scale. SN serves as the necessary catalyst—combining DE with SN produces the best SFT policy ($\mathbf{87.6}$ PDMS). Combined with Tab.~\ref{tab:exploration_diagnostics}, these results show that SN successfully converts diverse exploratory examples into actionable knowledge and provides an optimal initialization for downstream RL.

\noindent\textbf{Ablation on Diversity-Aware RL.}
We ablate our RL-stage training designs in Tab.~\ref{tab:ablation_rl_components}. The key challenge is filtering data to ensure GRPO receives non-zero advantages and meaningful gradients. We found that strategies inspired by difficulty-aware sampling (\eg, \cite{zhang2025grpolead}), such as our \textit{Human Difficulty} implementation, consistently lead to training collapse (35.2 PDMS), similar to \textit{Random Sample} and \textit{Full Trainset}. This suggests that avoiding zero-advantage scenarios is critical. We then explored filtering based on reward diversity, inspired by the ``medium-difficulty samples'' \cite{zheng2025greso}. A simple heuristic \textit{Reject Unimodal Distribution } Strategy successfully avoids collapse and achieves 88.8 PDMS. This confirms that filtering for reward diversity is a viable direction. Our proposed ADAS, which statistically validates this diversity via a Bernoulli test (see Sec.~\ref{subsec:darl}), performs even better (89.6 PDMS). Finally, combining the full ADAS (with 3 outer-loops) and Spanning Driving Reward (SDR) achieves the optimal 90.3 PDMS.

\begin{table}[h]
\centering
\small
\setlength{\tabcolsep}{3.5pt} 

\caption{Ablation study on Imitation Learning approaches. We evaluate the impact of three key Feasible Trajectory Expansion components: \textbf{DE} (Exploratory Data Expansion),\textbf{CoT} (Chain-of-Thought) and \textbf{SN} (Step-wise Normalization).}
\label{tab:ablation_il_components}
\begin{tabular}{ccc|ccccc|c}
\toprule
\textbf{DE} & \textbf{CoT} & \textbf{SN} & \textbf{NC}$\uparrow$ & \textbf{DAC}$\uparrow$ & \textbf{EP}$\uparrow$ & \textbf{TTC}$\uparrow$ & \textbf{C}$\uparrow$ & \textbf{PDMS}$\uparrow$ \\
\midrule
$\times$ & $\times$ & $\times$ & 97.7 & 91.8 & 85.8 & 96.8 & \textbf{98.4} & 83.9 \\
$\times$ & \checkmark & $\times$ & 98.2 & 93.2 & 85.8 & 97.3 & \textbf{98.4} & 85.6 \\
\midrule
\checkmark & \checkmark & $\times$ & 98.0 & 93.0 & 85.9 & 97.2 & \textbf{98.4} & 85.2 \\
$\times$ & \checkmark & \checkmark & 98.2 & 94.3 & \textbf{86.7} & 97.3 & \textbf{98.4} & 86.9 \\
\checkmark & \checkmark & \checkmark & \textbf{98.3} & \textbf{95.1} & 86.5 & \textbf{97.6} & 98.3 & \textbf{87.6} \\
\bottomrule
\end{tabular}
\end{table}

\begin{table}[h]
\centering
\footnotesize
\setlength{\tabcolsep}{3.5pt} 

\caption{Ablation study on RL approaches. We evaluate the effect of different \textbf{Sampling} strategies and the \textbf{SDR} (Spanning Driving Reward) on Navsim v1. The $i$x denotes a total of $i$ outer-loop iterations of ADAS.}
\label{tab:ablation_rl_components}
\scalebox{0.95}{
\begin{tabular}{lc|ccccc|c}
\toprule
\textbf{Sampling Strategy} & \textbf{SDR} & \textbf{NC}$\uparrow$ & \textbf{DAC}$\uparrow$ & \textbf{EP}$\uparrow$ & \textbf{TTC}$\uparrow$ & \textbf{C}$\uparrow$ & \textbf{PDMS}$\uparrow$ \\
\midrule
Human Difficulty & $\times$ & 73.9 & 43.7 & \textbf{94.9} & 70.3 & 97.1 & 35.2 \\
Reject Unimodal & $\times$ & \textbf{98.4} & 96.0 & 87.0 & 97.8 & \textbf{98.4} & 88.8 \\
ADAS(1x) & $\times$ & 98.2 & 96.3 & 88.6 & 97.6 & 98.2 & 89.6 \\
ADAS(3x) & $\times$ & \textbf{98.4} & 96.8 & 88.5 & 97.8 & 98.1 & 90.1 \\
\midrule
ADAS(3x) & $\checkmark$ & \textbf{98.4} & \textbf{96.9} & 88.5 & \textbf{97.9} & 98.1 & \textbf{90.3} \\
\bottomrule
\end{tabular}}
\end{table}

\section{Conclusion}

The ``exploit-explore'' dilemma represents a fundamental and persistent challenge in end-to-end autonomous driving systems. 
For VLA, although imitation learning (IL) provides a robust foundation by leveraging high-value ground-truth data that encapsulate the most probable driving behaviors, it still suffers from limited behavioral diversity due to data scarcity. Reinforcement Learning (RL) by itself, as the subsequent process of IL, is insufficient to solve the dilemma, particularly well-suited for critic-free RL algorithms (\eg GRPO) in VLMs.
Only through significantly enhanced diversity in both data and policy representations can we achieve more comprehensive and reliable planning capabilities in open-world environments.

In this paper, we first reveal and analyze the largely neglected Narrow Policy in Drive VLA due to the exploit-explore imbalance.
Accordingly, we introduce Curious-VLA, a systematic framework regarding data, sampling, and reward during both IL and RL and therefore balance exploitation and exploration, paving the way for more capable and reliable autonomous driving systems.

\section*{Acknowledgements}
This research was supported by the National Natural Science Foundation of China (Grant No. 62550184), Xiongan AI Institute, Lenovo Research and Wuxi Research Institute of Applied Technologies, Tsinghua University under Grant 20242001120.

{
    \small
    \bibliographystyle{ieeenat_fullname}
    \bibliography{main}
}

\clearpage
\setcounter{page}{1}
\maketitlesupplementary
\appendix

\section{Training Implement Details}

we provide more detailed configurations for both the Imitation Learning (SFT) and Reinforcement Learning (RL) stages.

\subsection{Training Stages on SFT}

Since the base Qwen2.5-VL model does not inherently support the specific \texttt{<think>} token, we introduce external tokens (\texttt{<thinking></thinking>} to wrap the \textit{driving explanation} part, \texttt{<answer></answer>} to wrap the \textit{trajectory prediction} part) to the tokenizer. To ensure the model effectively learns this structured reasoning format without compromising its visual encoding capabilities, we adopt a two-stage SFT strategy:

\begin{itemize}
    \item \textbf{Thinking Alignment.} In this stage, we freeze the vision encoder and the projector while optimizing only the LLM backbone. It primarily focuses on aligning the model with the structured CoT format and external tokens, ensuring the LLM captures the syntactic structure of the thinking process without disturbing the pretrained visual representations.

    \item \textbf{End-to-End Fine-tuning.} In the second stage, we unfreeze all parameters to end-to-end fine-tuning. This step enables joint optimization of the vision encoder and LLM, effectively bridging the vision-language space while adapting the pretrained knowledge to the domain shift of autonomous driving.

\end{itemize}

\subsection{Hyper-Parameters} 

We provide the detailed hyper-parameters for SFT in Table~\ref{tab:training_details_sft} 
and GRPO in Table~\ref{tab:training_details_grpo}.

\begin{table}[ht]
\centering
\caption{Training configurations for SFT.}
\label{tab:training_details_sft}
\begin{tabular}{ll}
\hline
Setting                 & SFT   \\
\hline
Base Model              & Qwen2.5-VL-3B   \\
Max Pixels              & 262144    \\
Global Batch Size       & 128        \\
Epochs                  & 3(align) / 3(fine-tune)      \\
Trainset Samples        & 103k(navtrain) + 39k(DE)         \\
Learning Rate           & 4e-5 / 5e-6       \\
Weight Decay            & 0.05       \\
Warmup Ratio            & 0.10       \\
bfloat16                & \checkmark \\
Global Batch Size       & 128         \\
GPUs                    & 8         \\
\hline
\end{tabular}
\end{table}

\begin{table}[ht]
\centering
\caption{Training configurations for GRPO.}
\label{tab:training_details_grpo}
\begin{tabular}{ll}
\hline
Setting                 & GRPO   \\
\hline
Max Pixels              & 262144    \\
Rollout Batch Size      & 256 \\
Actor Global Batch Size     & 256  \\
N Rollout(Group Size)       & 8     \\
Outer Loops                  & 3     \\
Total Steps       & 130         \\
Active Samples         & 6k/3k/1k (for 3 outer-loop)      \\
bfloat16                & \checkmark \\
GPUs                    & 8         \\
\hline
\end{tabular}
\end{table}

\subsection{Details on Span Driving Reward}

In Sec.~\ref{subsec:sdr}, we redesign the reward function by adapting the EPDMS into an additive focal objective with strict safety constraints. Furthermore, we remove $\text{EC}$ for training efficiency, and the sparse reward $R_{sparse}$ defined by Eq.~\ref{eq:r_sparse} uses $C' =\{\text{NC, DAC, DDC, TLC}\}$, and $M' = \{\text{EP, TTC, C, LK}\}$, the weights are $w'_m = \{5, 5, 2, 2\}$ and focal exponents are $\gamma_m = \{0.5, 0.5, 1.0, 1.0\}$.

\section{Data Pipeline}

\subsection{Details on Exploratory Data Expansion}

\noindent\textbf{Semantic-Aware Challenging Sceneario Filtering.}
We first identify 12k challenging driving segments (e.g., multi-lane roads, intersections, occlusions) from the 103k \texttt{navtrain} split. Although visual grounding models like Grounding-DINO can detect road elements, and the Navsim dataset itself provides rich SemanticMapLayers annotations, these sources cannot directly filter for challenging scenes that semantically possess ``multiple feasible trajectories.'' Therefore, we employ the following prompt with Qwen2.5-VL-72B to screen for these scenarios for data expansion. As shown in the prompt in Fig.~\ref{fig:filter_prompt}, the VLM is instructed to focus on complex road elements and identify scenarios that allow for diverse maneuvers.

\begin{figure*}[t]
    \centering
    \begin{tcolorbox}[title=\textbf{Prompt and Examples for Sceneario Filtering}, enhanced, fontupper=\small]
\textbf{Question:} 

\texttt{<image>} Give you a scene during driving. Observe the provided image to identify challenging scenarios, such as intersections, multi-lane roads, and occlusions.
Pay strict attention to the ego vehicle's current lane, ground markings, and traffic signs.
Provide your analysis strictly in the format below. Do not add any extra explanations. 
Always make conservative decisions. Unless the non-lane-keeping driving intention is very safe and necessary, only stay in the lane.

\textbf{Analysis Guidelines}
\begin{itemize}[leftmargin=*, nosep]
    \item An "intent" is an \textbf{immediately executable and traffic-compliant} high-level maneuver (e.g., turn left, change lane).
    \item Do \textbf{not} list intents that are clearly unreasonable or illegal (e.g., turning left from a straight-only lane, or changing lanes over a solid line).
\end{itemize}

\textbf{Output Format}
\begin{enumerate}
    \item SCENE\_SUMMARY: [Provide a brief (200-word max) description of the ego vehicle's situation.]
    \item ALL\_INTENTS: [List all plausible driving intents, separated by commas.]
    \item NON\_LANE\_KEEPING\_INTENT: [Based on the list in \#2, is there any valid intent other than lane following? Answer "Yes" or "No"]
\end{enumerate}

\noindent\dotfill\\
\textbf{Answer Example 1}:
\begin{enumerate}
    \item SCENE\_SUMMARY: The ego vehicle is on a two-lane road with a clear view ahead. The road is relatively empty, with no immediate traffic or pedestrians in sight. On the left side, there are trees, a sidewalk, and streetlights, while the right side has some vegetation and a \"No Parking\" sign. The road markings include a double yellow line in the center and a single yellow line on the right edge. The sky is partly cloudy, and a bridge is visible in the distance. The overall environment appears calm and safe for driving. There are no immediate obstacles or hazards on the road.
    \item ALL\_INTENTS: continue straight
    \item NON\_LANE\_KEEPING\_INTENT: No
\end{enumerate} 

\noindent\dotfill\\
\textbf{Answer Example 2}:
\begin{enumerate}
    \item SCENE\_SUMMARY: The ego vehicle is driving on a busy urban street, likely in a commercial area with multiple lanes and traffic signals. The vehicle is positioned in the right lane, following a red pickup truck. To the left, there is an orange van with advertisements, and to the right, there are various commercial buildings, including a Starbucks and a Planet 13 dispensary. The traffic signal ahead is red, indicating that vehicles must stop. The road markings are clear, with solid and dashed lines indicating lane boundaries. There are no immediate obstacles or pedestrians in the immediate vicinity of the ego vehicle. The overall environment suggests a typical city street with moderate traffic.
    \item ALL\_INTENTS: stop at the red light, change lane left, change lane right.
    \item NON\_LANE\_KEEPING\_INTENT: Yes
\end{enumerate}  

\end{tcolorbox}
    \caption{The prompt template and representative answer examples used to filter semantic-aware challenging scenarios with multiple feasible driving intents.}
    \label{fig:filter_prompt}
\end{figure*}

\noindent\textbf{Generative Trajectory Expansion.} 
To construct a robust exploratory dataset, we implement a rigorous expansion pipeline using the DDIM planner from ReCogDrive. To induce behavioral variance beyond deterministic outcomes, we modify the standard DDIM sampling by scaling the standard deviation of the Gaussian noise injected during the reverse denoising steps. The expansion process includes:

\begin{enumerate}
    \item \textbf{Hybrid Sampling Strategy:} We apply distinct sampling methods based on scene complexity. For the \textit{entire} \texttt{navtrain} dataset, we perform \textbf{intra-intent} ($k=32$) inference using the original prompts to explore execution variations within the same intent. For the 12k filtered challenging scenes, we additionally execute \textbf{inter-intent} inference by systematically altering the intention prompts (switching among \textit{Go Straight}, \textit{Turn Left}, \textit{Turn Right}, and \textit{Unknown}) to uncover plausible intents.
    
    \item \textbf{Safety and Diversity Filtering:} All generated candidates undergo a dual-criteria filter. For safety, a trajectory is retained only if its PDMS score exceeds $95.0$ and not less than the score of the human GT. For diversity, we employ a greedy selection based on geometric distance. We remove trajectories near the human GT and iteratively retain candidates, pruning those within a predefined distance margin.
\end{enumerate}

This pipeline ultimately yields \textbf{39k} non-ground-truth yet physically feasible exploratory samples, significantly broadening the policy's behavioral coverage beyond the narrow human ground truth.

\subsection{Details on Chain-of-Thought}

Following Poutine~\cite{rowe2025poutine}, we adopt a structured reasoning approach to explicitly model the decision-making process. The model processes the input through input context and four thinking tasks:

\noindent\textbf{Input Context.} The input $\mathcal{X}$ comprises a single \textit{CAMERA-FRONT} image, the ego-vehicle's kinematic history (past 1.5s trajectory, current velocity and acceleration), and the high-level driving intent.

\noindent\textbf{Thinking Tasks.} The CoT follows this sequence:
\begin{enumerate}
    \item \textbf{Critical Object Perception:} Identify the presence of critical objects~\cite{rowe2025poutine} that might influence the future path.
    \item \textbf{Driving Explanation:} Generate a concise, natural-language rationale (approximate 100 words).
    \item \textbf{Meta-Behavior Description:} Classify the intended behavior into discrete categories~\cite{rowe2025poutine}, specifically selecting the appropriate Speed and Command.
    \item \textbf{Trajectory Prediction:} Predict the optimal 4-second trajectory waypoints (normalized) conditioned on the previous steps.
\end{enumerate}.

\section{Inference Efficiency}

We evaluate real-time efficiency under serial inputs. As shown in Tab.~\ref{tab:suppl_latency}, Curious-VLA achieves 1.57s per sample, which is 7.74s faster than AutoVLA's text waypoint mode and competitive with its optimized Action+RFT mode (1.31s). The efficiency gain primarily benefits from our concise output template and single-view (1x C) input.

\begin{table}[t]
\centering
\footnotesize
\caption{Inference Latency Comparison. \textbf{Text}: Text waypoint; \textbf{Action}: Action token. AutoVLA uses a Fast-Slow Dual-System.}
\label{tab:suppl_latency}
\begin{tabular}{llc}
\toprule
\textbf{Method} & \textbf{Setting} & \textbf{Latency (s)} \\
\midrule
AutoVLA & Dual-Sys (Text) & 9.31 \\
AutoVLA & Dual-Sys (Action) & 3.95 \\
AutoVLA & Dual-Sys (Action + RFT) & \textbf{1.31} \\
\textbf{Curious-VLA} & Slow Think only (Text) & \underline{1.57} \\
\bottomrule
\end{tabular}
\end{table}

\section{RL Training Stability}

We provide the RL training curves to demonstrate the stability of our training pipeline. As shown in Fig.~\ref{fig:suppl_rl_curve}, we report Val Reward and Test PDMS over the entire 130 training steps (ADAS 3x, 3 outer-loops as described in Alg.~1). Outer-loop transitions are determined by Val Reward trends. In contrast, the \textit{Random Sample} baseline (without ADAS) shows RL collapse, confirming the necessity of diversity-aware sampling.

We further analyze training stability across multiple runs in Fig.~\ref{fig:suppl_rl_stability}. We perform $k=4$ independent training runs (ADAS 1x). The Critic (on trainset) and Val Reward curves demonstrate consistent improvement with low variance across all trials.

\section{External Analytical Experiments}
 We further extend the analysis to DiffusionDrive~\cite{liao2025diffusiondrive} in Tab.~\ref{tab:extension_diagnostics}. Despite its diverse 20-candidate pool(@all), the final Top-1 confidence-selected trajectory(@1) collapses to a single mode with the lowest diversity (0.037 / 0.076 mean-pADE/FDE), confirming that the narrow policy problem persists even in diffusion-based planners. In contrast, Curious-VLA achieves a superior balance across Quality, Diversity, and Performance.

\begin{table}[t]
\footnotesize
\centering
\setlength{\abovecaptionskip}{2pt}
\caption{Extended exploration analysis. DiffusionDrive is evaluated with its 20 denoised candidates(@all) and the confidence Top-1 selection(@1). All metrics @k=8.}
\label{tab:extension_diagnostics}
\begin{tabular}{llccc}
\toprule
\textbf{Method} & \textbf{Output} & \textbf{Quality} & \textbf{Diversity} & \textbf{Perf.} \\
\midrule
DiffusionDrive & Diff.@all & 0.218 / 0.430 & 0.571 / 1.175 & 87.60 \\
DiffusionDrive & Diff.@1 & 0.350 / 0.720 & 0.037 / 0.076 & 88.10 \\
\midrule
\textbf{Curious-VLA} & AR@1 & \textbf{0.269 / 0.547} & \textbf{0.641 / 1.415} & \textbf{91.55} \\
\bottomrule
\end{tabular}
\end{table}

\begin{figure}[ht]
\centering
    \setlength{\abovecaptionskip}{2pt}
    \includegraphics[width=0.8\linewidth]{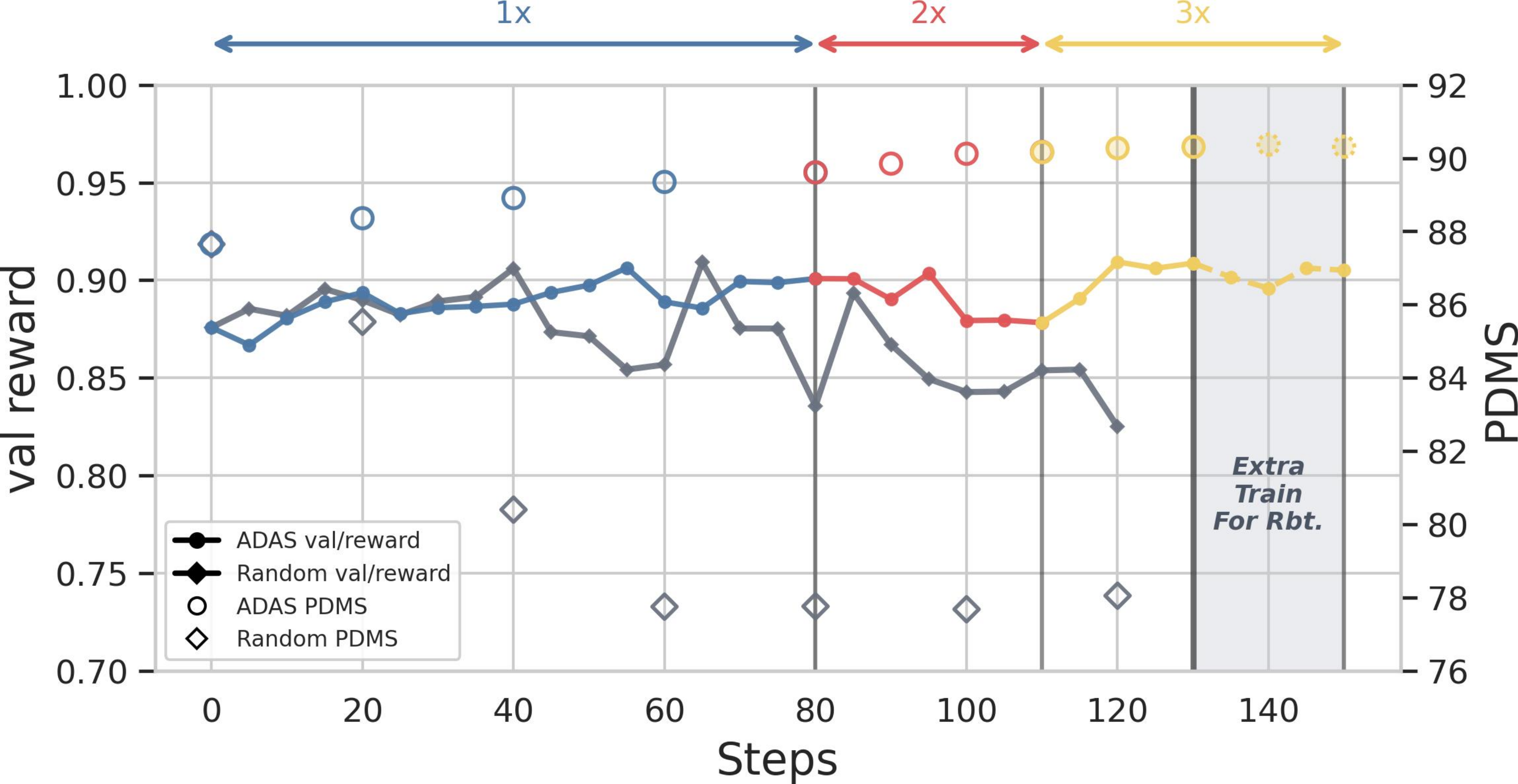}
    \caption{\textbf{RL Training Curves.} Val Reward and Test PDMS over 130 steps (ADAS 3x). The \textit{Random Sample} baseline shows RL collapse.}
    \label{fig:suppl_rl_curve}
\end{figure}

\begin{figure}[ht]
\centering
    \setlength{\abovecaptionskip}{2pt}
    \includegraphics[width=0.7\linewidth]{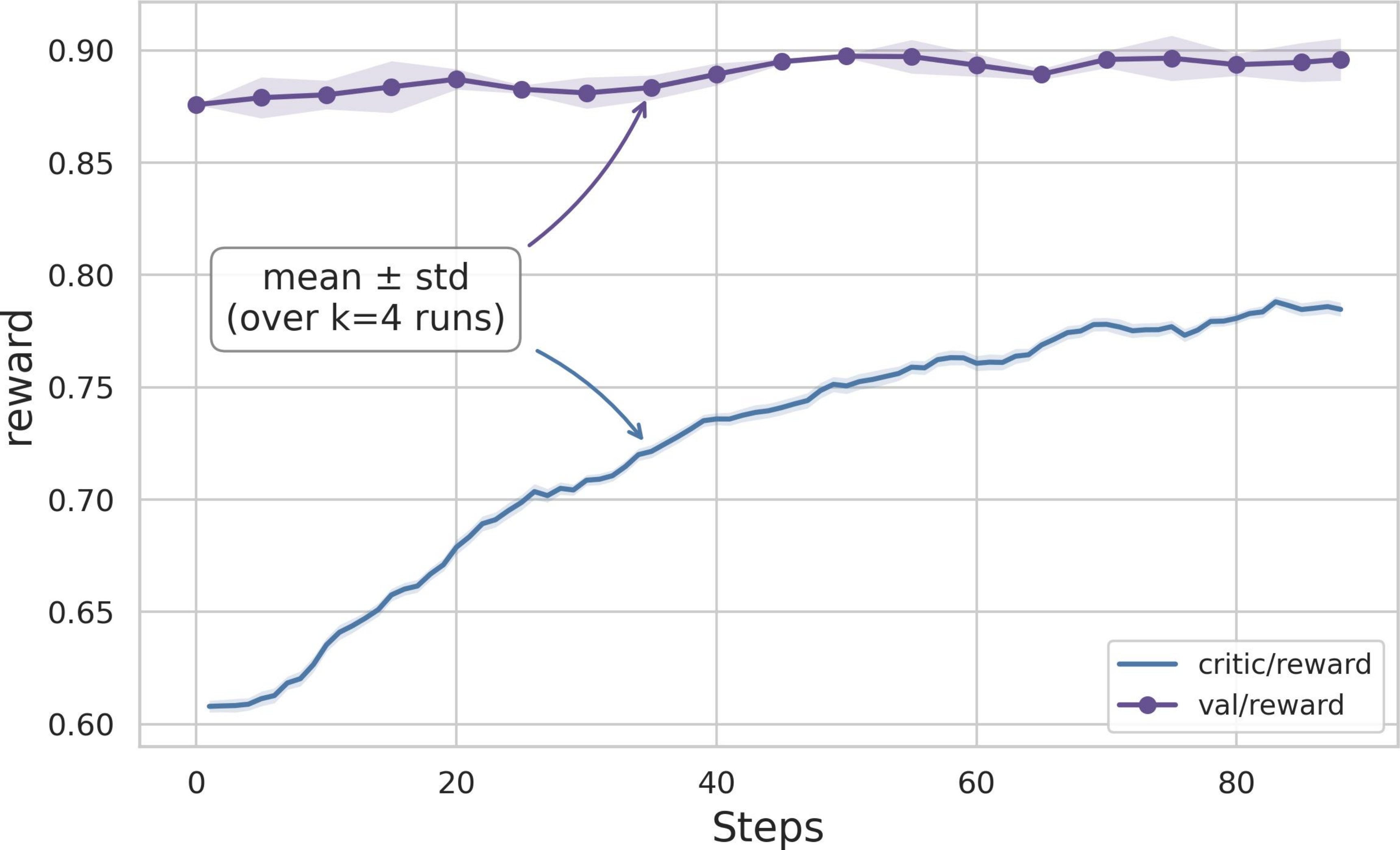}
    \caption{\textbf{Stability Analysis.} $k=4$ training runs (ADAS 1x). Critic and Val Reward curves show consistent improvement with low variance.}
    \label{fig:suppl_rl_stability}
\end{figure}

\section{More Visualization of Curious-VLA}

\begin{figure}[ht]
    \centering
    \includegraphics[width=0.8\linewidth]{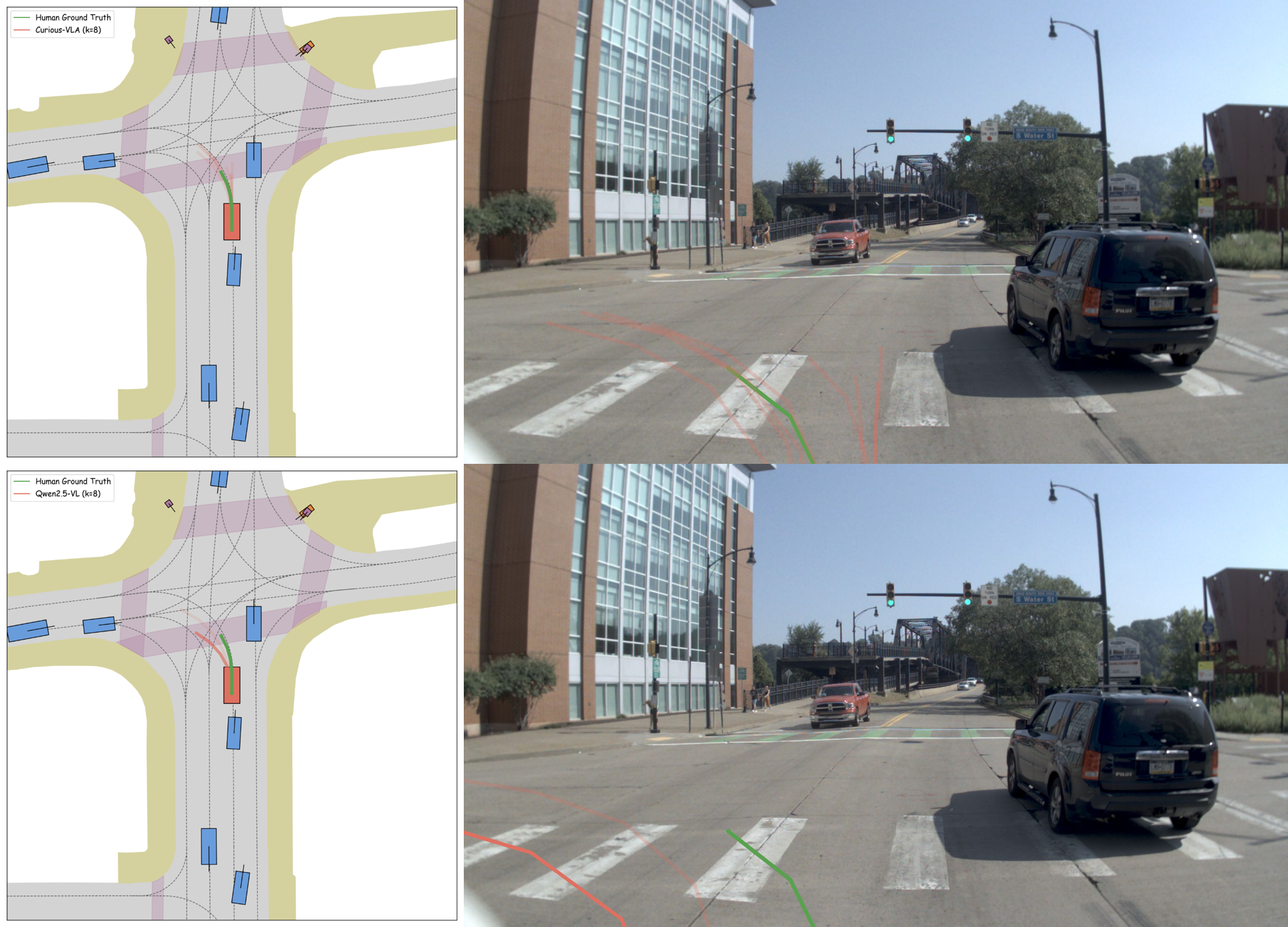}
    \includegraphics[width=0.8\linewidth]{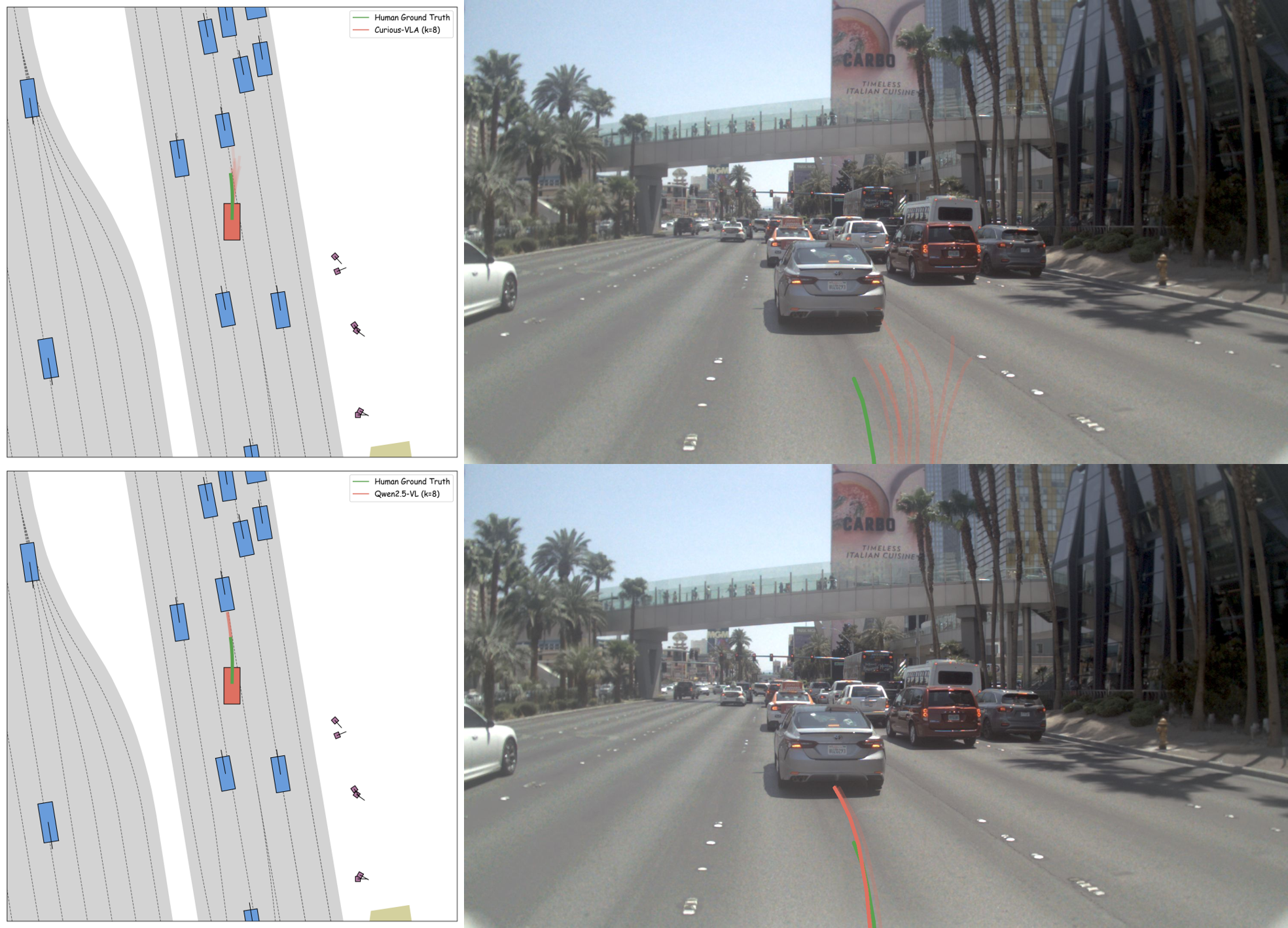}
    \includegraphics[width=0.8\linewidth]{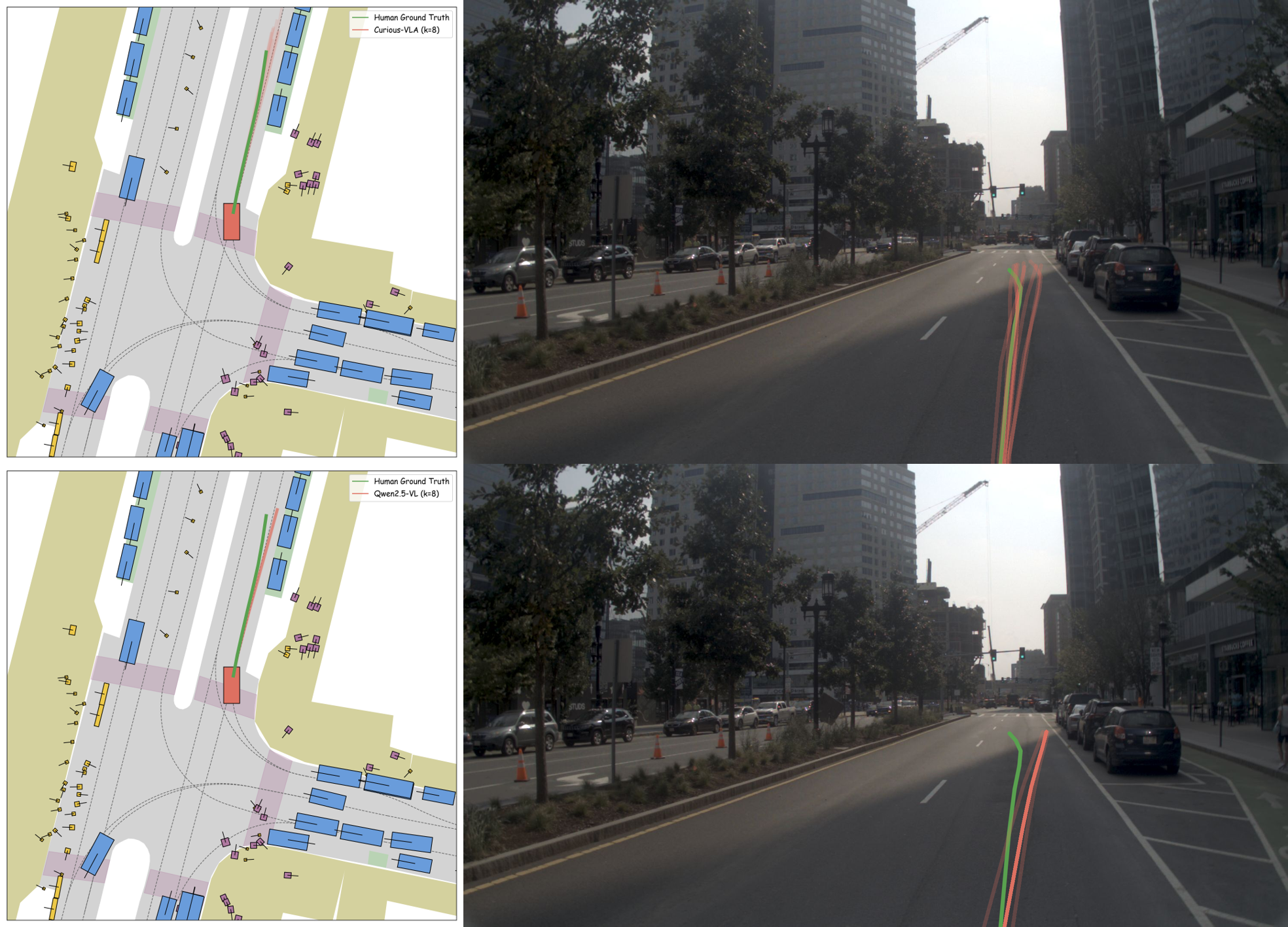}
    \caption{More visualization between Curious-VLA(top) and Qwen2.5-VL(bottom).}
    \label{fig:suppl_vis}
\end{figure}

As shown in Fig.~\ref{fig:suppl_vis}, \textbf{Curious-VLA} successfully alleviates the narrow policy bottleneck, ensuring diverse feasible behaviors.

\end{document}